\documentclass[a4paper,fleqn,10pt]{cas-dc}
\usepackage{color}
\usepackage{tikz}

\usepackage{graphicx}
\usepackage{caption}
\usepackage[authoryear]{natbib}
\usepackage{hyperref}
\usepackage{setspace}
\usepackage[switch]{lineno} 
\usepackage{amssymb} 

\hypersetup{
hidelinks,
colorlinks=true,
linkcolor=red,
citecolor=cyan,
urlcolor = black
}

\def\tsc#1{\csdef{#1}{\textsc{\lowercase{#1}}\xspace}}
\tsc{WGM}
\tsc{QE}
\tsc{EP}
\tsc{PMS}
\tsc{BEC}
\tsc{DE}

\begin{document}
\let\WriteBookmarks\relax
\def\floatpagepagefraction{1}
\def\textpagefraction{.001}



\title [mode = title]{DPF-Nutrition: Food Nutrition Estimation via Depth Prediction and Fusion}                      



%

\author[1]{\textcolor{black}{Yuzhe Han}}
\author[1]{\textcolor{black}{Qimin Cheng}}
\cormark[1]
\author[2]{\textcolor{black}{Wenjin Wu}}
\author[1]{\textcolor{black}{Ziyang Huang}}

\affiliation[1]{organization={School of Electronic Information and Communication, Huazhong University of Science and Technology},
    city={Wuhan},
    postcode={430074}, 
    country={China}}


\affiliation[2]{organization={Institute of Agricultural Products Processing and Nuclear Agricultural Technology, Hubei Academy of Agricultural Science},
    city={Wuhan},
    postcode={430064}, 
    country={China}}


\cortext[cor1]{Corresponding author}
\cortext[0]{E-mail address:\textcolor{cyan}{chengqm@hust.edu.cn} (Qimin Cheng).}


\begin{abstract}
A reasonable and balanced diet is essential for maintaining good health. With the advancements in deep learning, automated nutrition estimation method based on food images offers a promising solution for monitoring daily nutritional intake and promoting dietary health. While monocular image-based nutrition estimation is convenient, efficient, and economical, the challenge of limited accuracy remains a significant concern. To tackle this issue, we proposed DPF-Nutrition, an end-to-end nutrition estimation method using monocular images. In DPF-Nutrition, we introduced a depth prediction module to generate depth maps, thereby improving the accuracy of food portion estimation. Additionally, we designed an RGB-D fusion module that combined monocular images with the predicted depth information, resulting in better performance for nutrition estimation. To the best of our knowledge, this was the pioneering effort that integrated depth prediction and RGB-D fusion techniques in food nutrition estimation. Comprehensive experiments performed on Nutrition5k evaluated the effectiveness and efficiency of DPF-Nutrition.

\end{abstract}




\let\printorcid\relax
\maketitle

\section{Introduction}
Dietary health has become the predominant focus in modern life. Excessive and imbalance intake may lead to different kinds of diet-related diseases, especially obesity, which will dramatically increase the risk of  hypertension, cardiovascular and diabetes\citep{1}. Misestimating of nutrition content is a significant factor to excessive and imbalance intake. The International Food Information Council (IFIC) Foundation reported that that most people tend to overestimate their vegetable intake while underestimating their fat intake\citep{5}. Therefore, there is an urgent demand for effective nutrition estimation methods to help individuals  monitor their daily dietary intake and guide them to a healthier diet. Previous dietary assessment method heavily relied on human involvement. Specifically, the commonly used 24-hour Dietary Recalls\citep{6} requires participants report their food types and portion size over a 24-hour period, thereby understanding their eating behavior. Many popular applications, i.e. MyFitnessPal, MyDietCoach, Yazio, FatSecret, MyFoodDiary, and Foodnotes, are all developed based on this method. Despite of the advantage of easy to implement, it is burdensome and unreliable due to its high dependence on the subjective judgements of the participants. 

Fortunately, recent development in Artificial Intelligence (AI),
especially in deep learning techniques, have made the automated and reliable dietary assessment a reality. Vision-based nutrition estimation method allows users to monitor their food intake by capturing images using their mobile devices, which heavily reduces user burden. According to the type of input data,existing methods can be broadly divided into three  categories\citep{9}: monocular image-based methods, multi-view image-based methods and RGB-D methods.Earlier works \citep{7,8} have predominantly relied on multi-view images to reconstruct the 3D structure of food objects and estimate their volume. Subsequently, they calculated the sum of nutrition by combining the volume with the nutritional information of the food. Nevertheless, multi-view image methods are troublesome and inefficient since they require users to capture images from specific angles. In contrast, monocular image-based only relied on a single food image and demonstrated good performance.Specifically, \cite{10}proposed a method for estimating calories based on an energy density map, which maps RGB images to the energy density of food on a pixel-to-pixel basis. They compared the proposed method with the manual24-hour Dietary Recalls and the results showed an obvious advantage. Similarly,\cite{11}demonstrated that the performance of the vison-based method outperformed the professional human nutritionists  where they used a multi-task convolutional neural networks to estimate multiple nutrients. \cite{12}employed a combination of non-destructive detection technology and deep learning to analyze the nutritional content of food. They improved the detection of small target foods  to improve the nutrient estimation accuracy.  However, estimating food nutrition from a monocular image is an ill-posed problem\citep{2}. This is because the process of mapping food objects to monocular images often leads to a loss of crucial 3D details, which are vital for food portion estimation. In order to resolve the problem, depth information is utilized to complement the 3D information lost in monocular images.\cite{14} utilized RGB-D pairs captured from real eating scenarios as input and integrated techniques of food segmentation, recognition, and 3D surface reconstruction to estimate nutrient intake for hospitalized patients. \cite{11} incorporated the monocular images with depth maps as 4-channel data, which was subsequently sampled into a three-channel tensor as the input of model.\cite{13} improved their previous monocular image-based work\citep{10} through concatenating the features of energy density map and depth map at the last layer of the model.  \cite{15} proposed a nutrition estimation network integrating multi-scale and multi-modal fusion and achieved impressive results. This research demonstrated that the well-designed RGB-D fusion model outperformed the straightforward  fusion methods \citep{11,13} in incorporating RGB and depth images for nutrition estimation. Compared with monocular image-based methods, RGB-D methods have significant advantages in accuracy. However, the reliance of depth map acquisition on professional depth sensors increases the cost of the RGB-D methods and restricts the application scenarios.

In this paper,we proposed an estimation method called DPF-Nutrition,which combined the convenience and affordability of monocular image-based methods with the superior accuracy of RGB-D methods. Specifically, we employed a depth prediction model that  generated depth maps from monocular images to address the reliance on depth sensors for acquiring depth maps. These predicted depth maps recovered the missing 3D information in monocular images, thereby enhancing the precision of food portion estimation. Due to the distinct modalities of RGB and depth images, straightforward concatenating or summing cannot fully exploit the cross-modal features. To fully explore the complementary information for nutrition estimation, we designed a novel RGB-D fusion module integrating the proposed multi-scale fusion network and  cross-modal attention block (CAB). Within the RGB-D fusion module, multi-scale fusion network enhanced the semantic features through combining features of different resolutions, thereby preserving the intricate details of fine-grained food images that significantly contributed to nutrition estimation. The CAB eliminated redundancy and enhanced complementarity across RGB and depth features through employing an interactive attention mechanism, enabling the model to correctly focus on the nutrient regions. We evaluated the effectiveness of DPF-Nutrition on the public Nutrition5K dataset and obtained an encouraging results. Specifically, the percentage mean absolute errors (PMAE) for calorie, mass, protein, fat and carbohydrate estimation reached 14.7\%, 10.6\%, 20.2\%, 22.6\% and 20.7\%, respectively. With the mean PMAE of 17.8\%, our approach surpassed the results achieved by previous monocular image-based methods and remained competitive when compared to RGB-D methods. 

The contribution of this paper can be summarized as three aspects:
\begin{enumerate}[(1)]
\item We proposed a novel monocular image-based nutrition estimation method  based on Depth Prediction and Fusion, referred as DPF-Nutrition. It was the first attempt to incorporate depth prediction and RGB-D fusion techniques in vision-based nutrition estimation. 
\item We designed an RGB-D fusion module that incorporated the proposed multi-scale fusion network and cross-modal attention block (CAB) to fully exploit the informative image features for nutrition estimation. 
\item Our proposed DPF-Nutrition demonstrated  effectiveness in accurately estimating multiple nutrients, which has been evaluated on the public dataset Nutrition5k. 
\end{enumerate}

\section{Materials and methods}
\subsection{Dataset}
In this paper, we evaluated our DPF-Nutrition method on the Nutrition5K dataset\citep{11}, which is the largest publicly accessible food dataset containing comprehensive nutrient annotations. This dataset comprises 20k short videos and 3.5k RGB-D images captured by an Intel RealSense camera, involoving approximately 5k distinct food dishes. Each dish within the dataset includes detailed information such as ingredient names, quantities, and associated macronutrient data calculated using the reliable USDA Food and Nutrient Database \citep{16}. \textcolor{cyan}{Fig.~1} demonstrates examples of images in Nutrition5K dataset. As an image-based nutrition estimation method, our DPF-Nutrition was evaluated on the 3.5k food images from Nutrition5K dataset. The Nutrition5K dataset provides a predefined split, dividing the data into training and testing subsets with a ratio of 5:1. Despite the presence of a small number of erroneous images, as shown in \textcolor{cyan}{Fig.~2}, we opted to maintain the integrity of the dataset and prevent performing additional data cleaning to ensure consistency in the comparison of methods within the research. 
\begin{figure}[h]
	\centering
		\includegraphics[scale=0.75]{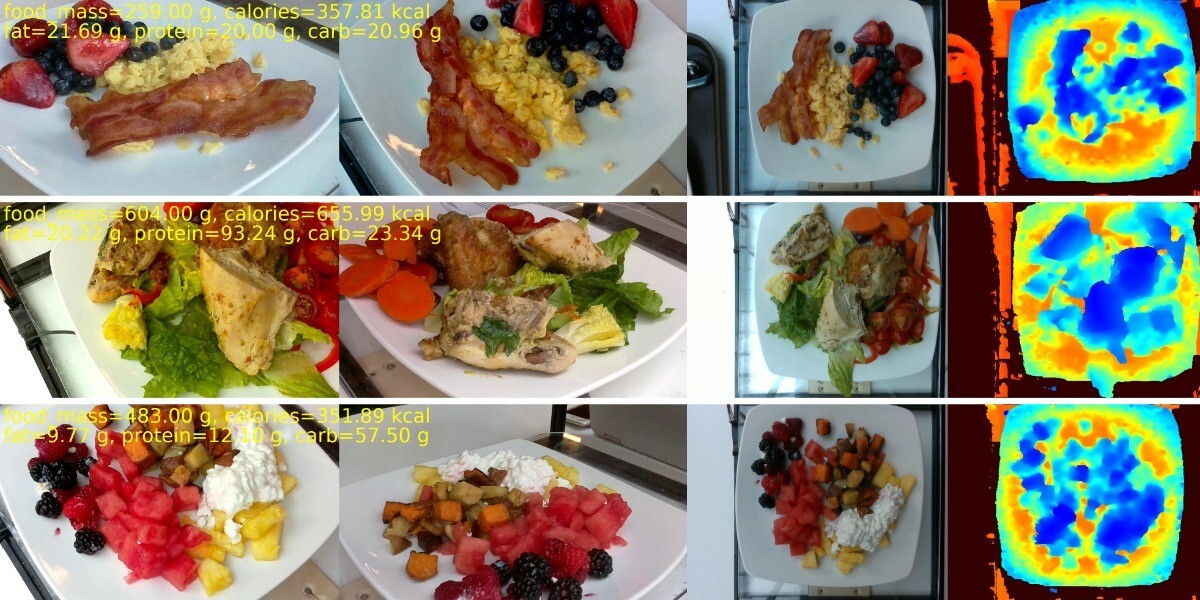}
	\caption{\textrm{The example images from Nutrition5k dataset.}}
	\label{FIG:1}
\end{figure}

\begin{figure}[h]
	\centering
		\includegraphics[scale=0.75]{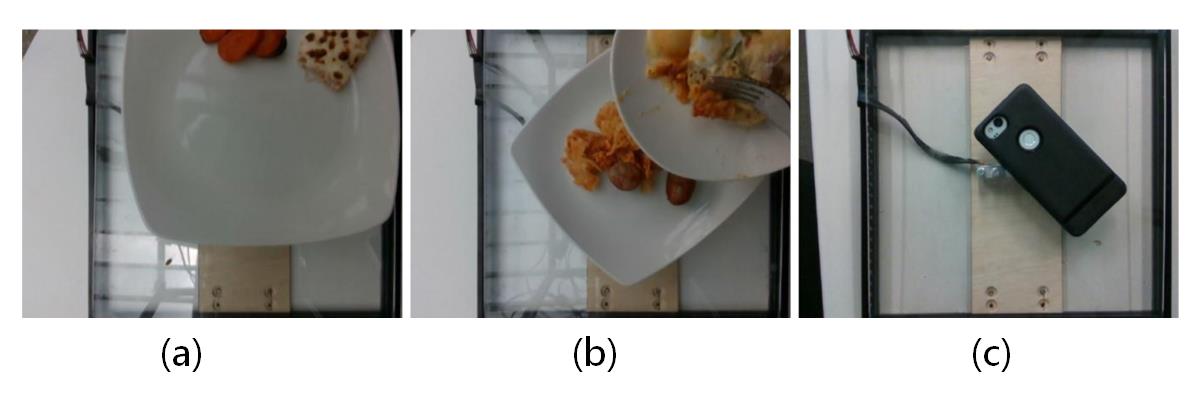}
	\caption{\textrm{Incorrect image samples. (a) food is not fully incorporated in the image. (b) dishes are overlapping. (c) non-food image.}}
	\label{FIG:2}
\end{figure}

\begin{figure*}
	\centering
		\includegraphics[scale=2]{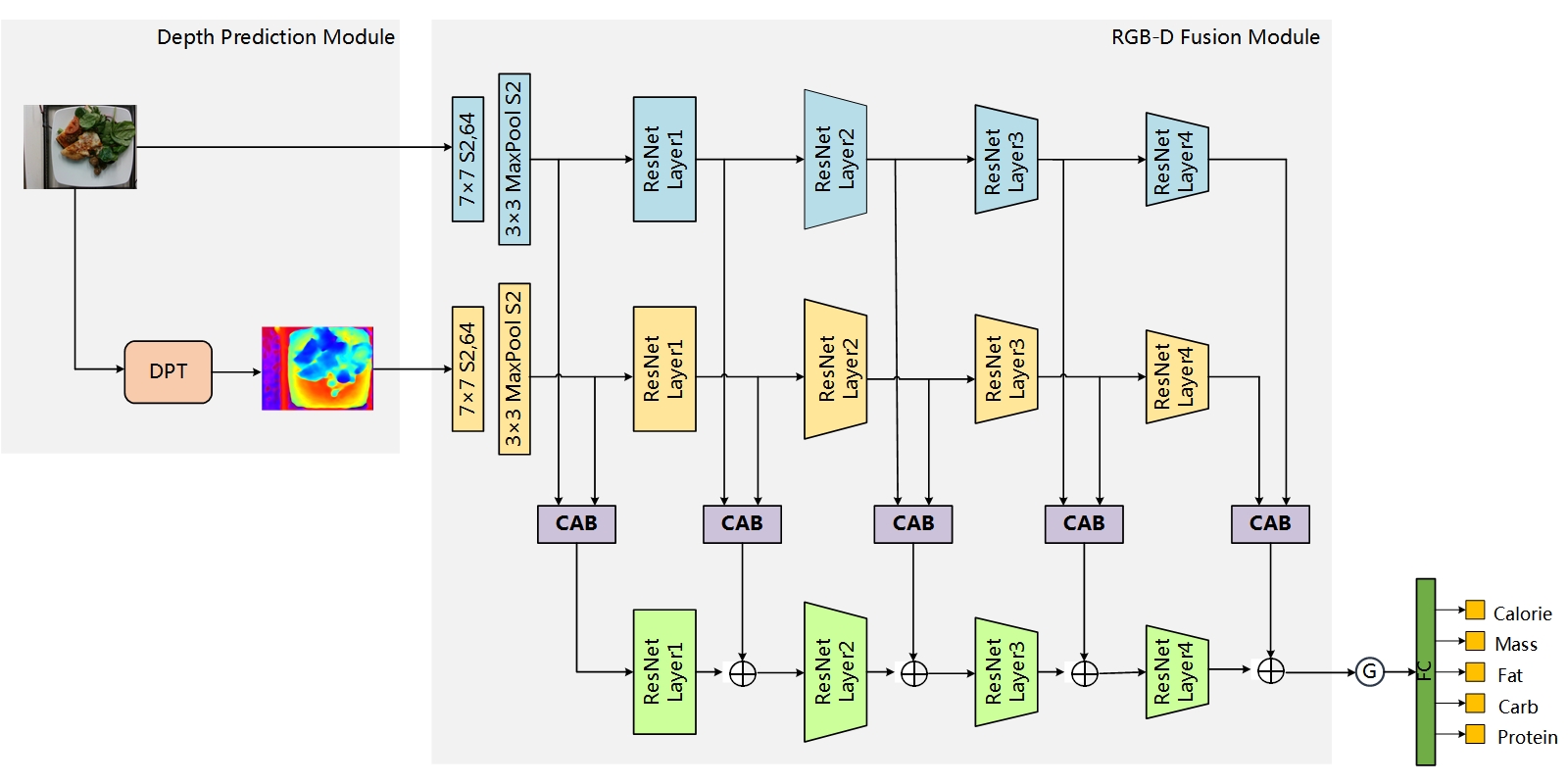}
	\caption{\textrm{The overall framework of our DPF-Nutrition, which consists of depth prediction module and RGB-D fusion module. We adopt depth prediction transformer(DPT) to generate the predicted depth map. We design a cross-modal attention block(CAB) to extract and integrate the complementary features of RGB and depth images. $\bigoplus$ indicates element-wise addition, $\large{\textcircled{\scriptsize{G}}}\normalsize$ denotes global average pool.}}
	\label{FIG:3}
\end{figure*}

\subsection{Methods}
\subsubsection{The DPF-Nutrition Structure}
Estimating nutrition through a monocular image is a challenging problem since the missing spatial information in monocular images is strongly related to food portion.
To tackle this problem, we proposed a monocular image-based nutrition estimation method based on Depth Prediction and Fusion, referred as DPF-Nutrition. The overall architecture of our DPF-Nutrition is illustrated in \textcolor{cyan}{Fig.~3}, comprising two modules:
\begin{enumerate}[(1)]
\item \textbf{Module1: Depth prediction module} aims to reconstruct the 3D depth information based on 2D monocular images. The depth prediction module employs vision transformer as the encoder which can reduce the loss of granularity and feature resolution and reconstruct the accurate depth information of food objects.
\item \textbf{Module2: RGB-D fusion module} is specifically designed to fully leverage the features of RGB and predicted depth images for accurate nutrition estimation. This module integrates a multi-scale fusion network and a cross-modal attention block (CAB). The multi-scale fusion network effectively enriches the intricate semantic features of fine-grained food images, while the CAB further enhances the complementarity of RGB and depth features.
\end{enumerate}

Specifically, we utilized the ResNet101 network\citep{24} as the backbone of our DPF-Net. For the depth prediction module, we employed the Dense Prediction Transformer (DPT)\citep{17} to generate a predicted depth map. This depth map was then combined with the RGB image to form the input for the RGB-D fusion module. We extracted RGB features and depth features using separate ResNet networks. The extracted features at the same ResNet layer were then fed into the cross-modal attention block (CAB). The CAB effectively suppressed redundant features within each modality and intelligently fused the complementary features. The fused RGB-D features of different resolutions were  combined through element-wise addition, starting from shallow layers and progressing to deep layers. This multi-scale fusion process ensured that the final feature map contained comprehensive and detailed information. The feature map was finally processed through global average pooling and a separate multi-task head to estimate the nutritional composition.

\subsubsection{The depth prediction module}
Depth prediction plays a crucial role in computer vision since it enhances the understanding and perception of real 3D scenes. The depth prediction model typically follows a pattern comprising an encoder and a decoder\citep{28,29,30}. The encoder extracts the features from the input images, while the decoder combines the features from the encoder and converts them into the ultimate depth prediction. The choice of the backbone network for the encoder is essential since the feature information that is lost in encoder cannot be recovered in following decoder. Compared with the depth information of other large scenes or objects, the depth information of food is more intricate due to its varied geometry and abundant texture. The loss of resolution and granularity during feature extraction in the encoder is more prone to causing distortion in the final depth prediction of food. However, the convolutional neural network\citep{24,25} inevitably loses granularity in deeper stages, since the increase in receptive field and abstraction of features are reliant on the downsampling operation. In comparison, Vision Transformer (ViT) \citep{18} abandons the downsampling operation and maintain the global receptive field of all stages. This makes it more suitable to be used as an encoder for fine-grained food images.

In this paper, we employed the Depth Prediction Transformer (DPT)\citep{17} to generate depth maps from monocular images. The Depth Prediction Transformer (DPT) comprised a transformer encoder and a convolutional decoder. The transformer encoder was responsible for extracting the bag-of-words representation, while the decoder reconstructed the bag-of-words representation into image-like features of different scales. Finally, image-like features were combined into the depth estimation. The structure of DPT is showed in \textcolor{cyan}{Fig.~4}. 

\begin{figure*}
	\centering
		\includegraphics[scale=1.4]{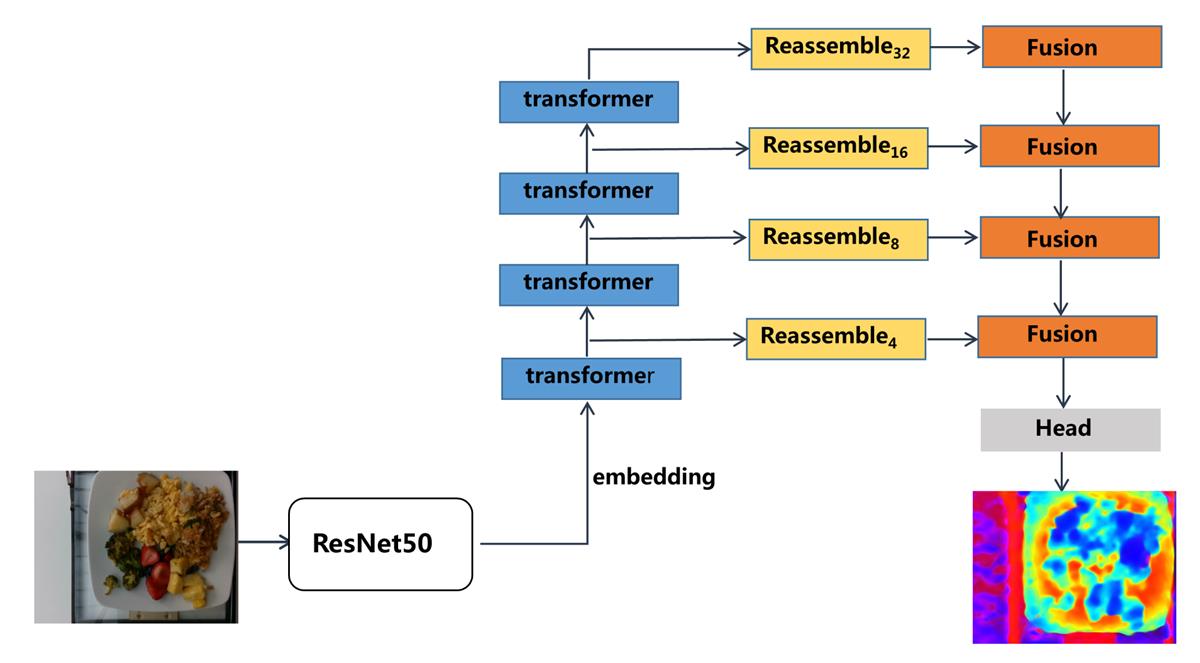}
	\caption{\textrm{The structure of depth prediction module. The input image is transformed into feature vectors by ResNet-50 feature extractor and consequently embedded into two-dimensional tokens. The tokens are then fed into transformer encoder. The tokens from different transformer stages are reassembled into image-like feature maps at various resolutions. Finally, the image-like feature maps are fused progressively to generate the depth prediction.}}
	\label{FIG:4}
\end{figure*}

Specially, the input images was abstracted as a two-dimensional feature vectors by ResNet-50 network to satisfy the input format of transformer. The feature vectors were then combined with a trainable position embedding to incorporate the positional information. transformer encoder, given that the resolution of the input image is H×W, we obtained a set of tokens t = \{$t_{0}$$\cdots$$t_{N_p}$\},$t_{n}$ $\in$ $R^{D}$,where $N_{p}$ = $\frac{H\times W}{p^{2}}$, D refers to the dimension of tokens, p is the sampling rate of ResNet50. Then the NP tokens were spatially concatenated into an image-like feature map through placing each token based on the information of the position embedding.

Then the $N_{p}$ tokens were spatially concatenated into an image-like feature map through placing each token based on the information of the position embedding:
\begin{equation}
    Concatnate: R^{N_{p}\times D} \rightarrow R^{\frac{H}{p}\times \frac{W}{p} \times D}
\end{equation}

We resampled the image-like representations into specific size $\frac{H}{s}\times \frac{W}{s}$ with $D^{'}$ dimensions:
\begin{equation}
    Resample: R^{\frac{H}{p}\times \frac{W}{p} \times D} \rightarrow R^{\frac{H}{s}\times \frac{W}{s} \times D^{'}}
\end{equation}

Finally, we employed a RefineNet-based decoder\citep{31} to progressively combine the feature maps at different resolutions and generate the depth prediction. More detailed implementation of DPT can be obtained in the paper\citep{17}.

\subsubsection{RGB-D fusion module}
RGB-D fusion is widely applied in various visual tasks including image classification\citep{19}, food intake detection\citep{20} and food nutrition assessment\citep{15} . The essence of RGB-D fusion is to combine the distinct information from RGB and depth images  to create enhanced and more informative features. Since RGB and depth images come from different modalities, a straightforward concatenation or summation approach is insufficient to fully exploit the valuable complementary information they possess. Nevertheless, many previous RGB-D nutrition estimation methods have employed these fusion methods without considering the inherent differences between modalities. For instance, incorporating the depth information as the additional 4th channel of the input images\citep{11}, or concatenating the features extracted from RGB and depth images at the final layer of the model \citep{13}.

Different from previous approaches, our proposed method introduced a novel cross-modal attention block (CAB) for RGB-D fusion to explore the distinction and complementarity of the cross-modal features. By utilizing the CAB, we enhanced informative features and filter out redundant ones, ensuring that only the most relevant information is retained before the fusion process. Unlike traditional attention mechanisms that typically focus on a single modality \citep{32,33}, our CAB leverages RGB-D interactive attention to enhance features. This  enables our model to accurately prioritize regions with high nutritional value, rather than solely focusing on specific food items or larger portions. Additionally, we incorporated a multi-scale fusion network to enhance the semantic representation of fine-grained food images. This enabled our model to focus more on intricate details and small food objects, resulting in improved nutrition estimation.

Specially, after the depth prediction module, the RGB images together with estimated depth images were input into the feature extraction networks to generate hierarchical features, denoted as RGB features \{R$_i$\}$_{i=0}^4$and depth features \{D$_i$\}$_{i=0}^4$. The RGB and depth feature maps, with the same resolution, were fed into CAB to generate the complementary fused representation. The structure of CAB was demonstrated in \textcolor{cyan}{Fig.~5}. The CAB explicitly built the feature correspondences among different modalities based on the channel and spatial attention vectors of additive features, which emphasized crucial features and suppress redundancy ones. The re-calibrated features were concatenated to obtain complementary cross-modality features. Specially, given two input features R$_i$ and D$_i$, we first added the two features pixel by pixel along the channel dimension. Then, we processed the additive features in two branches to obtain channel attention vectors and spatial attention vectors. In channel attention branch, we employed global average pool to obtain global channel descriptor. Then the global channel descriptor was fed into a 1×1 convolution with BN and ReLU to increase non-linearity. Finally, it through a sigmoid activation function to produce the channel attention vector, the procedure can be defined as:

\begin{equation}
    CA = Sig(Conv_{1\times 1}(GAP(R_i\oplus D_i)))
\end{equation}
where GAP($\cdot$) denotes the global average pooling, $\bigoplus$ denotes the element-wise addition, Sig($\cdot$) represents the sigmoid function, $Conv_{1\times 1}$  indicates a convolutional layer with 1×1 kernel size, followed by BN and ReLU.

In spatial attention branch, we calculated the average value for all pixels of the additive feature map along the channel dimension to obtain spatial descriptor. Then we applied a 3×3 convolution with BN and ReLU to smooth the spatial descriptor. Finally, the spatial attention vector was obtained by passing it through a sigmoid activation function, the procedure can be defined as:
\begin{equation}
    SA = Sig(Conv_{3\times 3}(Mean(R_i\oplus D_i)))
\end{equation}
where Mean($\cdot$)denotes mean function along the channel dimension.
The attention weights of channel and spatial dimensions enhanced the correlation and complementarity of the RGB and depth features. Based on the cross-modal attention, the enhanced RGB feature map and depth feature map were obtained. The enhanced features were then further concatenated and fed into a 1×1 convolution to obtain the complementary RGB-D features C$_i$, the procedure can be defined as:
\begin{equation}
    C_i = Conv_{1\times 1}(Concat(R_i\otimes CA \otimes SA, D_i\otimes CA \otimes SA ))
\end{equation}
where Concat($\cdot$) denotes the cross-channel concatenation, $\bigotimes$ indicates pixel-wise multiplication.

\begin{figure}[h]
	\centering
		\includegraphics[scale=1]{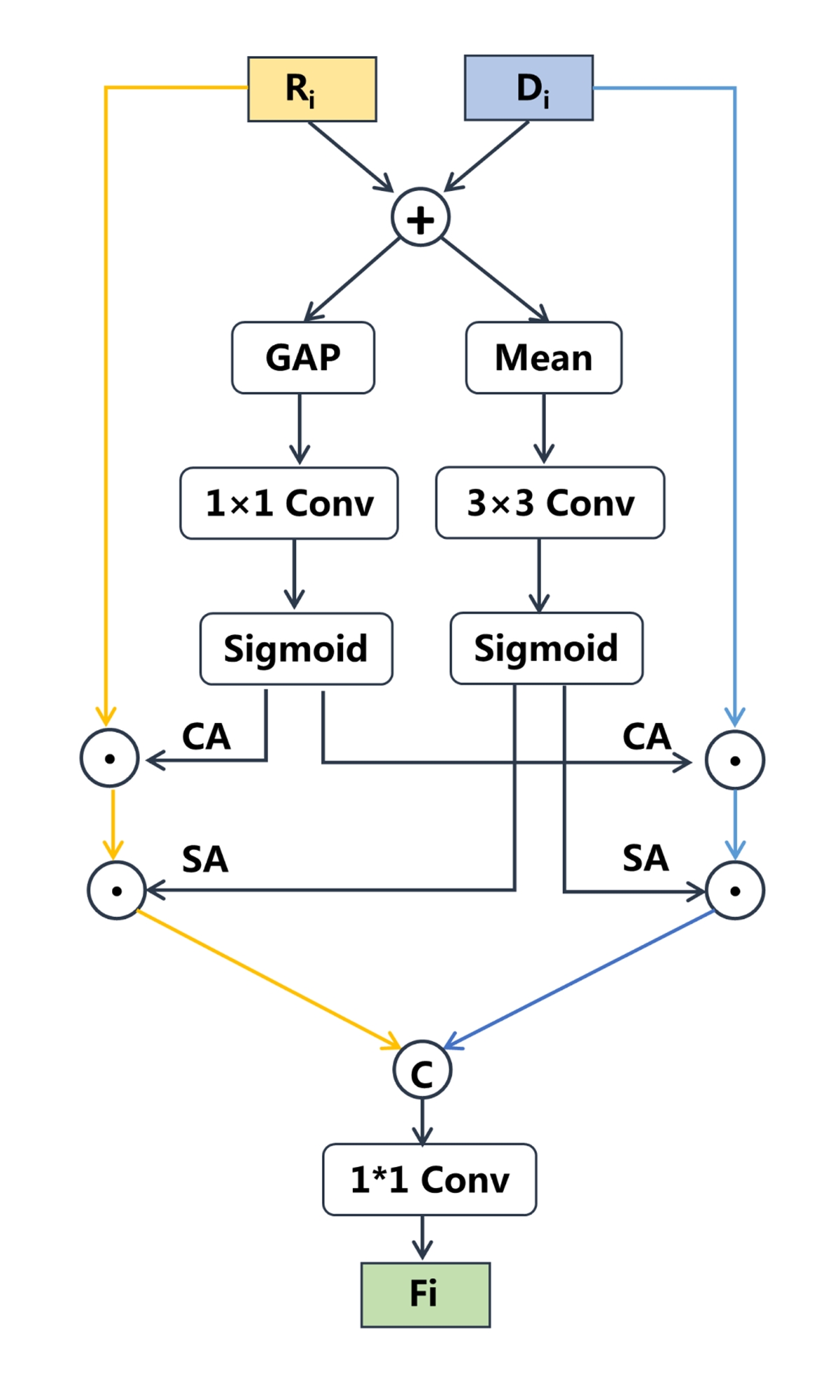}
	\caption{\textrm{The structure of CAB. GAP indicates global average pooling. $\bigoplus$ denotes the element-wise addition, $\bigotimes$ indicates pixel-wise multiplication. Mean represents mean function along the channel dimension.}}
	\label{FIG:5}
\end{figure}

After the proposed CAB, we obtained the cross-modal features \{C$_i$\}$_{i=0}^4$. To further enhance to semantic feathers, we adopted a multi-scale fusion network to combine the local detailed information of low-level features and the global context information of the high-level feathers, progressively.
Specially, we used the first cross-modal feature map F$_0$ as the input of the feature fusion network, the feature was then refined by a ResNet convolutional block and combined with the next cross-modal feature map C$_1$ to generate the fused feature map F$_1$. We repeated this operation at the next stage. In this way, we combined the cross-modal features at different scales. The procedure can be defined as:
\begin{equation}
    F_i = C_i \oplus Res_i(C_{i-1})
\end{equation}
where Res$_i$ indicates the i-th ResNet convolutional block.
Through the multi-scale fusion network and CAB, we fully utilized the complementary RGB and depth information to generate the final feature representation F$_4$. Then it was fed into a full connected (FC) layer and five multi-task FC heads (with dimension 2048 and 1,respectively) to generate the estimated nutrition values.

\subsubsection{Loss function}
As a multi-task learning model, our DPF-Nutrition predicts the contents of calorie, mass and three essential macronutrients of food. For each subtask, L1 loss was used to measure the bias between the estimated nutritional values and ground-truth ones, which can be defined as:
\begin{equation}
    L_{cal} = \frac{1}{N} \sum_{i=1}^{N}  \left| y_{cal}-y_{cal}^{'}  \right|
\end{equation}
where L$_cal$ denotes the subtask loss of calorie, y$_cal$ indicates the estimated value of calorie and y$_{cal}^{'}$ represents the ground-truth calorie value. The loss of other subtasks follow this equation. 

The scale of the subtask losses are various, which can cause that some tasks dominant other tasks during the training phase. The geometric loss combination \citep{21} is invariant to the scale of individual losses, thereby maintaining a balanced approach towards subtask losses of varying scales. Thus, We used the geometric loss strategy as our loss function, which can be defined as:
\begin{equation}
    L_{total} = \sqrt[5]{L_{cal}L_{mass}L_{fat}L_{carb}L_{protein}}
\end{equation}
where L$_{tatal}$ denotes the overall loss function.

\subsection{Evaluation metrics}
In this paper, we adopted two evaluation metrics of mean absolute error (MAE) and percentage of mean absolute error (PMAE), which are defined as:
\begin{equation}
   \rm MAE = \frac{1}{N} \sum_{i=1}^{N} \left| y_{i}-y_{i}^{'}  \right|
\end{equation}
\begin{equation}
   \rm PMAE = \frac{\rm MAE}{\frac{1}{N} \sum_{i=1}^{N} y_i} 
\end{equation}

where y is the estimated nutrient value while y$^{'}$  is the ground-truth nutrient value. Caloric values are measured in standard kilocalories units, while the other three nutrient values are measured in grams. A higher level of accuracy in nutrient estimation is achieved when the MAE and PMAE values for evaluation are lower.

\section{Results}
\subsection{Experimental detail}
 All the experiments were conducted on a 24G NVIDIA GTX 3090 GPU. To maintain the experimental fairness, the same setup was utilized for all experiments. For the training of depth prediction module, we resized the input images to have a long side of 384 pixels and train on random square crops of size 384. The encoder network was initialized with ImageNet1K pre-trained weight while the decoder network was initialized randomly. We utilized the Adam optimizer \citep{22} with an initial learning rate of 1e$^{-5}$ and implemented a cosine annealing strategy for learning rate decay. The learning rate was declined to 1e$^{-6}$ after cosine annealing. We trained the model for 60 epochs with a batch size of 8. For the training of RGB-D fusion module, the input images were resized to 336×448 pixels. Image augmentation methods including center cropped and random horizontal flip were applied to training images. The backbone network was initialized with Food2K\citep{23} pre-trained weight. We chose Adam optimizer with an initial learning rate of 5e$^{-5}$, and implemented an exponential decay strategy for updating the learning rate, with a decay rate set to 0.98. We trained the model for 150 epochs with a batch size of 8.

\subsection{Backbone comparison}
It is essential to select a suitable backbone networks for the model. In this section, we made a comparison among several widely used Convolutional Neural Networks (CNNs) and the recently popular vision transformer\citep{18}. According to \textcolor{cyan}{Table.~1}, ResNet101 achieved better performance, with the best mean PMAE of 20.9\%. Therefore, we used ResNet101\citep{24} as the backbone network for our DPF-Nutrition, unless otherwise specified.

\begin{table}[width=1\linewidth,pos=h]\scriptsize
\rmfamily 
\setlength\tabcolsep{3pt} 
\caption{\textrm{Comparison of the performance of different backbones.} }\label{tab1}
\begin{tabular*}{\tblwidth}{@{} Lcccccc@{} }

\toprule
Methods &
  \multicolumn{1}{c}{\begin{tabular}[c]{@{}c@{}}Calorie\\ \tiny{PMAE(\%)}\end{tabular}} &
  \multicolumn{1}{c}{\begin{tabular}[c]{@{}c@{}}Mass\\ \tiny{PMAE(\%)}\end{tabular}} &
  \multicolumn{1}{c}{\begin{tabular}[c]{@{}c@{}}Fat\\ \tiny{PMAE(\%)}\end{tabular}} &
  \multicolumn{1}{c}{\begin{tabular}[c]{@{}c@{}}Carb\\ \tiny{PMAE(\%)}\end{tabular}} &
  \multicolumn{1}{c}{\begin{tabular}[c]{@{}c@{}}Protein\\ \tiny{PMAE(\%)}\end{tabular}} &
\multicolumn{1}{c}{\begin{tabular}[c]{@{}c@{}}Mean\\ \tiny{PMAE(\%)}\end{tabular}}\\

\midrule
\begin{tabular}[c]{@{}l@{}}ViT\\ \tiny{\citep{18}}\end{tabular}        & 20.4          & 16.3         & 29.4         & 28.9         & 28.9         & 24.6         \\
\begin{tabular}[c]{@{}l@{}}VGG16\\ \tiny{\citep{25}}\end{tabular}     & 18.6        & 14.6         & 28.0         & 26.2        & 26.8          & 22.8         \\
\begin{tabular}[c]{@{}l@{}}InceptionV3\\ \tiny{\citep{26}}\end{tabular}  & 18.0         & 14.2          & \textbf{25.1} & 25.8         & 24.8         & 21.6         \\
\begin{tabular}[c]{@{}l@{}}ResNet50\\ \tiny{\citep{24}}\end{tabular}    & 18.2         & 13.8         & 28.0          & \textbf{22.8} & 25.9          & 21.7          \\
\begin{tabular}[c]{@{}l@{}}ResNet101\\ \tiny{\citep{24}}\end{tabular}    & \textbf{17.9} & \textbf{13.6} & 26.5          & 23.0          & \textbf{24.5} & \textbf{21.1} \\ 
\bottomrule
\end{tabular*}

\end{table}

\subsection{Method comparison}
To evaluate the performance of DFP-Nutrition, we compared it with other vision-based nutrition estimation methods. To ensure a fair comparison, we consistently maintained the experimental settings and dataset splits across all methods. The experimental results were showed in \textcolor{cyan}{Table.~2}. We referred the methods proposed by \cite{11} and \cite{15} as Google-Nutrition and RGB-D Nutrition, respectively. We compared different methods including: Google-Nutrition, RGB-D Nutrition, Swin-Nutrition\citep{12} and our proposed DPF-Nutrition. It should be noted that \cite{12} conducted additional data cleaning for their original experiments. To ensure fairness in our comparison, we replicated their method on our data. Consequently, the results of Swin-Nutrition in this paper differed from the original ones.

Google-Nutrition was introduced in conjunction with the publication of the Nutrition5K dataset, which was widely regarded as the benchmark method for nutrition estimation. Google-Nutrition offered two variations: Google-Nutrition-monocular, which utilized monocular images, and Google-Nutrition-depth, which incorporated depth data for nutrition estimation. To the best of our knowledge, both Swin-Nutrition and RGB-D Nutrition  achieved state-of-the-art performance in nutrition estimation. Swin-Nutrition focused on utilizing monocular images, while RGB-D Nutrition  demonstrated superior performance when using RGB-D images. As demonstrated by the experimental results, DPF-Nutrition outperformed other methods in terms of mean PMAE of nutrients. Compared to the monocular image-based Swin-Nutrition, DPF-Nutrition showcased a remarkable improvement. Specifically, the mean PMAE  improved by 2.6\%, while the MAE of calorie and mass achieved improvements of 3.6kCal and 6.3g, respectively. Furthermore, when compared to RGB-D Nutrition, DPF-Nutrition demonstrated a competitive performance with a 0.7\% improvement in the mean PMAE. The results proved the effectiveness of our DPF-Nutrition approach.

\begin{table*}\footnotesize
\rmfamily 
\caption{\textrm{Comparison of the performance of different methods. }}\label{tab2}

\setlength\tabcolsep{7pt}
\begin{tabular*}{\textwidth}{@{} llcccccc@{} }

\toprule
Input &
  Methods &
  \multicolumn{1}{c}{\begin{tabular}[c]{@{}c@{}}Calorie\\ \scriptsize{MAE / PMAE}\end{tabular}} &
  \multicolumn{1}{c}{\begin{tabular}[c]{@{}c@{}}Mass\\ \scriptsize{MAE / PMAE}\end{tabular}} &
  \multicolumn{1}{c}{\begin{tabular}[c]{@{}c@{}}Fat\\ \scriptsize{MAE / PMAE}\end{tabular}} &
  \multicolumn{1}{c}{\begin{tabular}[c]{@{}c@{}}Carb\\ \scriptsize{MAE / PMAE}\end{tabular}} &
  \multicolumn{1}{c}{\begin{tabular}[c]{@{}c@{}}Protein\\ \scriptsize{MAE / PMAE}\end{tabular}} &
\multicolumn{1}{c}{\begin{tabular}[c]{@{}c@{}}Mean\\ \scriptsize{PMAE}\end{tabular}}\\

  
\midrule
\multirow{1.8}{*}{RGB-D images} &
  \begin{tabular}[c]{@{}l@{}}Google-Nutrition-depth\\ \citep{11}\end{tabular} &
  47.6 / 18.8\% &
  40.7 / 18.9\% &
  \textbf{2.27 / 18.1\%} &
  4.6 / 23.8\% &
  3.7 / 20.9\% &
  20.1\% \\
 &
  \begin{tabular}[c]{@{}l@{}}RGB-D Nutrition\\ \citep{15}\end{tabular} &
  38.5 / 15.0\% &
  21.6 / 10.8\% &
  3.0 / 23.5\% &
  4.43 / 22.4\% &
  3.69 / 21.0\% &
  18.5\% \\
\midrule

\multirow{3}{*}{Monocular images} &
  \begin{tabular}[c]{@{}l@{}}Google-Nutrition-monocular\\ \citep{11}\end{tabular} &
  70.6 / 26.1\% &
  40.4 / 18.8\% &
  5.0 / 34.2\% &
  6.1 / 31.9\% &
  5.5 / 29.5\% &
  29.1\% \\
 &
   \begin{tabular}[c]{@{}l@{}}Swin-Nutritiom\\ \citep{12}\end{tabular}&
  41.5 / 16.2\% &
  27.5 / 13.7\% &
  3.21 / 24.9\% &
  4.32 / 21.8\% &
  4.47 / 25.4\% &
  20.4\% \\
 &
  DPF-Nutrition(ours)&
  \textbf{37.9 / 14.7\%} &
  \textbf{21.2 / 10.6\%} &
  2.92 / 22.6\% &
  \textbf{4.09 / 20.7\%} &
  \textbf{3.56 / 20.2\%} &
  \textbf{17.8\%} \\ 
\bottomrule
\end{tabular*}

\end{table*}

\subsection{Ablation study}
To validate the effectiveness of the proposed modules, comprehensive ablation studies were conducted on the various components comprising our DPF-Nutrition. The experimental results were illustrated in \textcolor{cyan}{Table.~3}. The baseline was the RGB stream that used only a ResNet101 network to estimate nutrients from RGB images. Similarly, depth stream used a single network to estimate nutrients from depth images generated by the depth prediction module. Model (c) and Model (d) employed simple feature vector concatenation and the proposed multi-scale fusion network, respectively, to integrate the RGB stream and depth stream. Model(e) represented the complete DPF-Nutrition which incorporates the CAB on the basis of model(d).

According to \textcolor{cyan}{Table.~3}, Model (a) achieved a mean PMAE of 20.9\%, whereas model (b) achieves a mean PMAE of 38.6\%. The results demonstrated that relying solely on depth information alone cannot accurately estimate nutrients. Model (c) concatenated the  RGB and depth features at the last layer of the feather extraction network and achieves a mean PMAE of 19.0\%, which is a 2.1\% improvement compared to Model (a). The result demonstrated the effectiveness of complementing RGB images with estimated images for nutrition estimation and validated the efficacy of our depth prediction module. The improvements achieved by Model (d) and Model (e)  evaluated the effectiveness of the proposed multi-scale fusion network and Cross-modal Attention Block (CAB), respectively. Model (e) achieved an improvement over Model (c), with a 1.2\% decrease in mean PMAE, as well as reductions of 2.8kCal in calorie MAE and 1.5g in mass MAE. The results of the ablation experiment demonstrated that all the depth prediction module, multi-scale fusion network, and CAB can effectively improve the accuracy of nutrient prediction.

\begin{table*}\footnotesize
\caption{\textrm{Comparison with different ablation settings. }}\label{tab3}
\rmfamily 
\setlength\tabcolsep{10pt}
\begin{tabular*}{\textwidth}{@{} llcccccc@{} }
\toprule
Index &
  Model &
   \multicolumn{1}{c}{\begin{tabular}[c]{@{}c@{}}Calorie\\ \scriptsize{MAE / PMAE}\end{tabular}} &
  \multicolumn{1}{c}{\begin{tabular}[c]{@{}c@{}}Mass\\ \scriptsize{MAE / PMAE}\end{tabular}} &
  \multicolumn{1}{c}{\begin{tabular}[c]{@{}c@{}}Fat\\ \scriptsize{MAE / PMAE}\end{tabular}} &
  \multicolumn{1}{c}{\begin{tabular}[c]{@{}c@{}}Carb\\ \scriptsize{MAE / PMAE}\end{tabular}} &
  \multicolumn{1}{c}{\begin{tabular}[c]{@{}c@{}}Protein\\ \scriptsize{MAE / PMAE}\end{tabular}} &
\multicolumn{1}{c}{\begin{tabular}[c]{@{}c@{}}Mean\\ \scriptsize{PMAE}\end{tabular}}\\
  
\midrule
(a)   & RGB Stream                   & 46.0 / 17.9\%   & 27.2 / 13.6\% & 3.42 / 26.5\% &4.56 / 23.0\%&	4.31 / 24.5\%	& 21.1\% \\
(b)   & Depth Stream                 &83.5 / 32.5\%	&44.7 / 22.3\%	&6.29 / 48.8\%	&8.08 / 40.8\%	&8.53 / 48.5\%	&38.6\% \\
(c)   & (a)+(b)+direct fusion       &40.7 / 15.8\%	&22.7 / 11.3\%	&3.17 / 24.6\%	&4.25 / 21.5\%	&3.87 / 22.0\%  	&19.0\% \\
(d)   & (a)+(b)+multi-scale fusion &39.3 / 15.3\%	&21.6 / 10.8\%	&3.01 / 23.3\%	&4.13 / 20.9\%	&3.72 / 21.2\%	&18.3\% \\
(e)   & (d) + CAB                  &\textbf{37.9 / 14.7\%}	&\textbf{21.2 / 10.6\%}	&\textbf{2.92 / 22.6\%}	&\textbf{4.09 / 20.7\%}	&\textbf{3.56 / 20.2\%}	&\textbf{17.8\%} \\
\bottomrule
\end{tabular*}

\end{table*}

\subsection{Visualization analysis}
To visually showcased the efficacy of our method, we begined by visualizing the performance of the depth prediction module. As depicted in \textcolor{cyan}{Fig.~6}, our depth prediction module successfully recovered the depth information and  smoothed out the noise in the original depth. Next, we employed Grad-Cam\citep{27} to visualize the RGB-D fusion module, which generated a heat map highlighting the regions of interest (ROI) identified by the model. This visualization allowed for a more intuitive understanding of how our method leveraged food images to make accurate predictions. The visualization results of certain dishes were showed in \textcolor{cyan}{Fig.~7}, and the corresponding nutrition facts were illustrated in \textcolor{cyan}{Table.~4}. As illustrated in the visualization, our model demonstrated a focused attention on specific regions for each nutrient task. For instance, when estimating the fat and protein content in dish\_156278816, the model exhibited a strong focus on the rice and corn components. Conversely, when estimating the carbohydrate content, the model primarily emphasized the rice and corn elements. This indicated that our model effectively captured relevant visual cues and assigned appropriate importance to different regions based on the specific nutrient being estimated.

\begin{figure}
	\centering
		\includegraphics[scale=0.75]{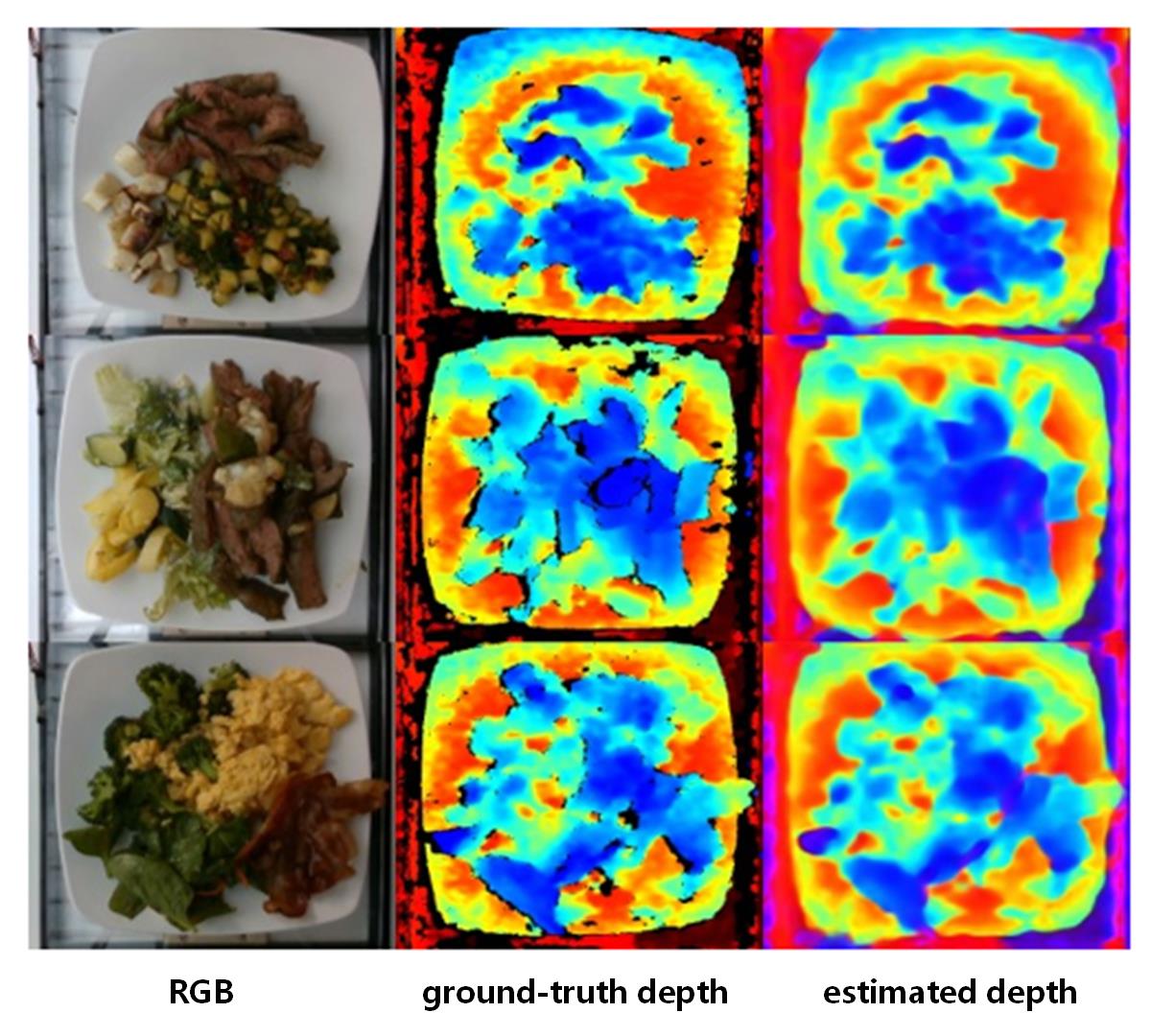}
	\caption{\textrm{The sample results of the depth estimation.}}
	\label{FIG:6}
\end{figure}

\begin{figure*}
	\centering
		\includegraphics[scale=1.8]{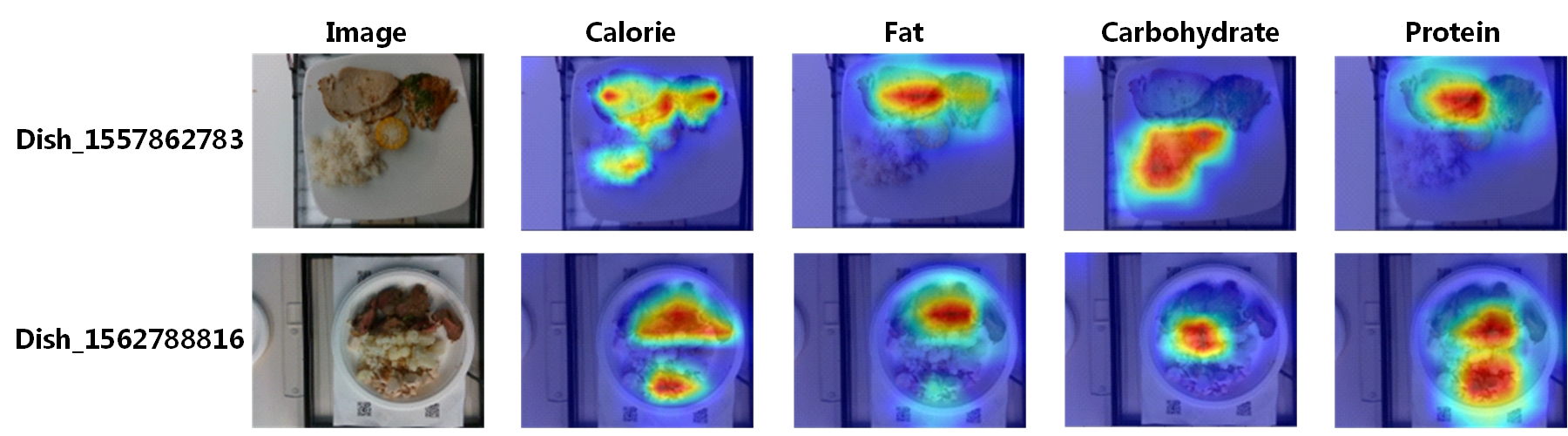}
	\caption{\textrm{The ROI heat-maps of different nutrients}}
	\label{FIG:7}
\end{figure*}

\begin{table}[width=1\linewidth,cols=6,pos=h]\footnotesize
\rmfamily 
\setlength\tabcolsep{9pt}
\caption{\textrm{The nutrition facts of the example dishes.} }\label{tab4}

\begin{tabular*}{\tblwidth}{@{} lcccc@{} }
\toprule
\multicolumn{5}{c}{Dish\_1562788816}                        \\
\midrule
Ingredient  & Calorie(kCal) & Fat(g) & Carb(g) & Protein(g) \\
Pork        & 231           & 13.6   & 0       & 25.3       \\
Corn        & 41            & 0.94   & 8.6     & 1.2        \\
Rice        & 133           & 0.31   & 28.8    & 2.8        \\
Fish        & 36            & 5.3    & 0       & 7.5        \\
  
\midrule
\multicolumn{5}{c}{Dish\_1557862783}                        \\
\midrule
Ingredient  & Calorie(kCal) & Fat(g) & Carb(g) & Protein(g) \\
Steak       & 320           & 22.4   & 0       & 29.5       \\
Chicken     & 104           & 2.3    & 0       & 19.5       \\
Cauliflower & 19            & 0.23   & 3.9     & 1.4       \\
\bottomrule
\end{tabular*}

\end{table}

\section{Discussion}
DPF-Nutrition demonstrated effective performance in estimating the essential nutritional intake in daily life. We used the model to estimate the content of calorie, mass, protein, fat and carbohydrate from a monocular image, and the errors were 14.7\%, 10.6\%, 20.2\%, 22.6\% and 20.7\%, respectively. In order to investigate the limitations of our model, we examined the test samples that yielded poor results.

We discovered that the most significant errors primarily originated from three categories of images.The example images were showed in \textcolor{cyan}{Fig.~8}.The first category includes images with food stacking and covering. For instance, in the case of dish\_1560367980, where the low-calorie spinach significantly obscured the high-calorie pizza, the estimated calorie value was 31.6\% lower than the actual value.The second category comprises images containing minuscule components that are imperceptible, such as oil and sugar. For example, in the case of dish\_1562617939, where a large amount of olive oil was added to the dish, the calorie PMAE was up to 52.5\%, while the fat PMAE was dramatically 68.3\%.The third category comprises images with food items that are rarely encountered in the training data. For example, in the case of dish\_1562617703, which involved the uncommon ice-cream in the Nutrition5k dataset, the calorie PMAE reached up to 39.5\%, while the carbohydrate PMAE reached up to 57.4\%. It is evident that food stacking, imperceptible components,  insufficient data are the primary factors that restrict the performance of DPF-Nutrition.Wherein the poor performance of the model on the images existing food stacking and imperceptible components exposed the limitation of computer vision technique. In addition, the success of deep learning technique relies heavily on data, however, the construction of the food dataset is costly and labor-intensive. As a result,the scarcity of food datasets with nutritional information has impeded the further progress of vision-based nutrition estimation.

In the future, we will focus on creating a large public food dataset for comprehensive dietary assessment.Furthermore, we are dedicated to overcome the limitations of vision-based methods for more accurate nutrition estimation.

\begin{figure}[h]
	\centering
		\includegraphics[scale=0.25]{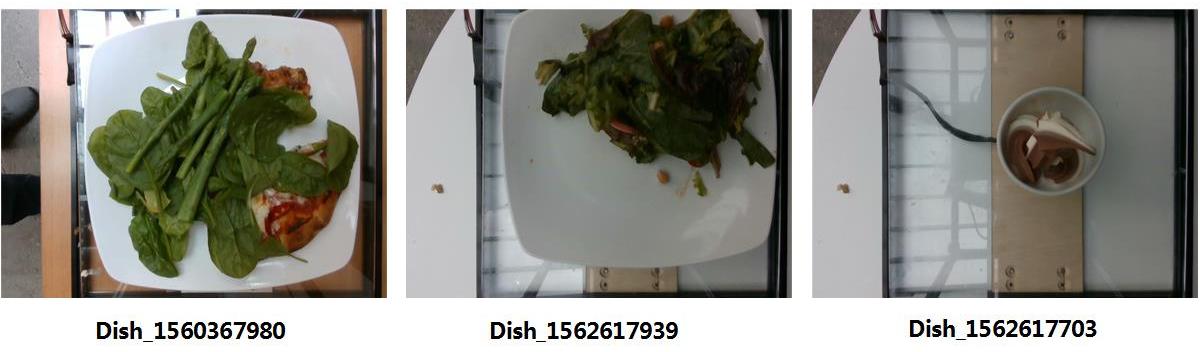}
	\caption{\textrm{Example of the samples with poor results.}}
	\label{FIG:8}
\end{figure}

\section{Conclusions }
In this paper, we proposed our DFP-Nutrition for dietary assessment, aiming to develop an automated, cost-effective, and precise method for nutrition estimation. Our proposed method offered a novel approach by predicting the depth map from a monocular image and incorporating the recovered 3D information with RGB food images to improve nutrition estimation. To assess the effectiveness of our method, we performed experiments on the Nutrition5K dataset and compared its performance with that of state-of-the-art image-based nutrition estimation methods. The results demonstrated the superiority of our proposed method over existing monocular image-based approaches. Moreover, when compared to RGB-D methods, our method showcased competitive performance, further solidifying its effectiveness. In the future, we envision the widespread utilization of automated vision-based nutrition estimation methods in our daily lives, making a significant contribution to improving people's dietary health.

\printcredits

\bibliographystyle{cas-model2-names}

\bibliography{reference}

\end{document}